%% file: main.tex
\documentclass[10pt,twocolumn,letterpaper]{article}

\usepackage[pagenumbers]{cvpr}

\usepackage{tabularx}
\usepackage{booktabs}
\usepackage{multirow}

\input{preamble}

\definecolor{cvprblue}{rgb}{0.21,0.49,0.74}
\usepackage[
    pagebackref,
    breaklinks,
    colorlinks,
    allcolors=cvprblue
]{hyperref}

\def\confName{CVPR}
\def\confYear{2027}

\title{Making Future-Predictive Representations Actionable}

\author{
    Fangchuan Wang\textsuperscript{1}\quad
    Hao Zhang\textsuperscript{2}\quad
    Qiyan Zhao\textsuperscript{3}\quad
    Huanqi Zhao\textsuperscript{1}
    \\[5pt]
    Guanting Ye\textsuperscript{4}\quad
    Xu-Yao Zhang\textsuperscript{3,*}\quad
    Yuqing Gao\textsuperscript{2,*}
    \\[10pt]
    {\small
    \textsuperscript{1}Xinjiang University
    \qquad
    \textsuperscript{2}Tongji University
    }
    \\[3pt]
    {\small
    \textsuperscript{3}Institute of Automation,
    Chinese Academy of Sciences
    }
    \\[3pt]
    {\small
    \textsuperscript{4}University of Science and Technology of China
    }
    \\[7pt]
    {\footnotesize
    \textsuperscript{*}Corresponding authors
    }
}

\begin{document}

\raggedbottom
\maketitle


\begin{figure*}[t]
    \centering
    \includegraphics[width=\textwidth]{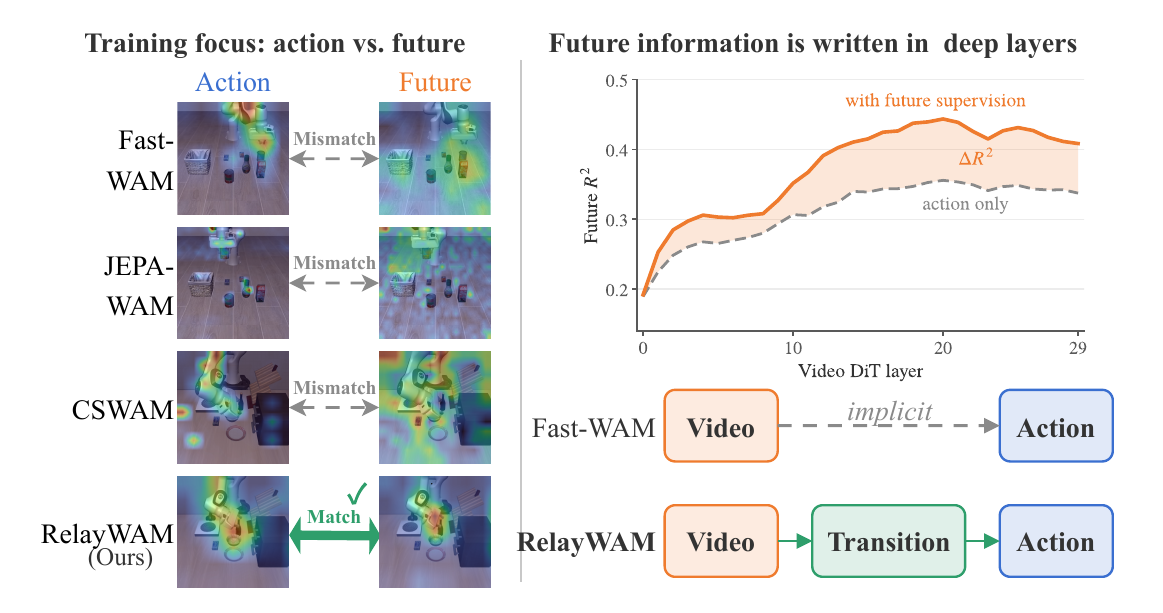}
    \caption{
        \textbf{Overview of RelayWAM and the representation--utilization gap.}
        \textbf{Left:} Training-focus attribution reveals that existing future-supervised WAMs suffer from an objective mismatch—action relies on local gripper--object regions while future prediction attends to the whole scene—whereas RelayWAM aligns both focuses.
        \textbf{Top right:} Linear probing shows that future supervision writes rich predictive cues ($\Delta R^2$) into deeper Video DiT layers.
        \textbf{Bottom right:} RelayWAM introduces an explicit transition space to relay these cues to the Action DiT without test-time future generation, reaching 72.03\% success on LIBERO-Plus.
    }
    \label{fig:teaser}
\end{figure*}


\input{sec/0_abstract}
\input{sec/1_intro}
\input{sec/2_relatedwork}
\input{sec/3_Analysis}
\input{sec/4_method}
\input{sec/5_experiments}


\section{Conclusion}
\label{sec:conclusion}

We uncover a \emph{representation--utilization gap} in future-free inference World Action Models: while future supervision writes rich predictive information into deep visual layers, it lies along directions largely orthogonal to those read by the action pathway.
To bridge this gap, we propose \textbf{RelayWAM}, whose recurrent transition space extracts task- and state-conditioned predictive cues and injects them into native action attention, outperforming even models that decode future frames at inference.
Advancing world action models requires not only accurate future prediction, but also explicit, action-aligned interfaces that let the action pathway read it.


\input{sec/6_llm_statement}


{
    \small
    \bibliographystyle{ieeenat_fullname}
    \bibliography{main}
}


\input{sec/X_suppl}

\end{document}

%% file: preamble.tex
\usepackage{cuted} 

\usepackage{currfile} 

\usepackage{caption} 

\usepackage{amsmath}
\usepackage{amssymb}
\usepackage{graphicx}
\usepackage{xcolor}
\usepackage{booktabs}
\usepackage{threeparttable}
\usepackage{multirow}
\usepackage{colortbl}
\usepackage{svg}
\usepackage[most]{tcolorbox}
\definecolor{relaynavy}{RGB}{60,84,136}
\definecolor{relaylight}{RGB}{221,227,239}

\newtcolorbox{highlightbox}{
  enhanced,
  colback=relaylight,
  colframe=relaynavy,
  boxrule=0.6pt,
  arc=3pt,
  left=5pt, right=5pt, top=2pt, bottom=2pt,
  before skip=4pt, after skip=4pt,
  fontupper=\small
}

\usepackage{booktabs}
\usepackage{multirow}
\usepackage{colortbl}
\definecolor{fstbg}{RGB}{186,210,232}   
\definecolor{sndbg}{RGB}{225,236,245}   
\newcommand{\fst}[1]{\cellcolor{fstbg}\textbf{#1}}
\newcommand{\snd}[1]{\cellcolor{sndbg}#1}
\usepackage{pifont}

%% file: sec/0_abstract.tex
\begin{abstract}
Recent world action models (WAMs) drop future prediction at inference, assuming that representations learned through auxiliary future supervision still benefit action generation.
However, our layerwise analysis of Fast-WAM reveals a depth-dependent mismatch: while probes decode future state transitions more accurately from deeper visual layers, the gradients and visual focuses of the future and action objectives increasingly diverge with depth, leaving much of this predictive information unused by the action pathway.
To address this issue, we introduce \textbf{RelayWAM}, which adds an explicit transition space between the video and action backbones.
Conditioned on the task instruction, robot state, and current action estimate, recurrent transition layers extract step-relevant predictive cues across depth and relay them directly into the action attention, without generating future frames at inference.
Experiments show that RelayWAM substantially amplifies the benefit of future supervision, translating into stronger generalization under distribution shifts in both simulation and real-world robot tasks.
On LIBERO-Plus, RelayWAM achieves \textbf{72.03\%} mean success, compared to 56.42\% without future supervision and 51.50\% for Fast-WAM.
\end{abstract}

%% file: sec/1_intro.tex
\section{Introduction}
\label{sec:intro}

World action models (WAMs) bring future prediction into action learning.
Built on large-scale pretrained video models, they jointly learn to generate actions and to predict future observations~\cite{motus,cosmospolicy,lingbotva,videovla,dreamzero,gr1,vpp,uwm,uva,genieenvisioner,gigaworldpolicy,nativevap,trajwam}.
Recent work goes further: future-prediction supervision improves policy performance even when the future-generation branch is removed at inference, and future prediction serves only as an auxiliary training objective~\cite{fastwam,flare,vlajepa,jepawam,lightwam}.
The intuition is straightforward.
A model that can predict ``what happens next'' should encode object dynamics and future state changes, and this should help action generation generalize.

This intuition, however, rests on an implicit assumption.
In Fast-WAM~\cite{fastwam} and similar models, future supervision can improve action generation through the shared representation.
This design assumes that the action pathway actually uses the information that future supervision writes into this representation.
Existing methods are validated mainly by success rates in simulation and on real robots.
These results show that future supervision helps, but not how the representations it shapes are used.
This leaves an open question:

\begin{highlightbox}
\textcolor{relaynavy}{\textbf{Question.}} What representations does future supervision shape, and are they actually used by action generation?
\end{highlightbox}

To answer this question, we analyze future-supervised WAMs from two sides, \emph{writing} and \emph{reading}.
On the writing side, we introduce training-focus attribution, which reveals the visual content that each objective actually relies on.
We find that the two objectives are only partly aligned.
Future information is indeed written into deeper layers, while the gradients of the two objectives become less aligned and their training-focus maps gradually drift apart.
On the reading side, we erase this information from the features read by the action pathway.
The resulting action change is close to that caused by erasing random directions of the same rank.
These findings reveal a discrepancy between the information shaped by future supervision and the information utilized by action generation.

\begin{figure}[t]
  \centering
  \includegraphics[width=\linewidth]{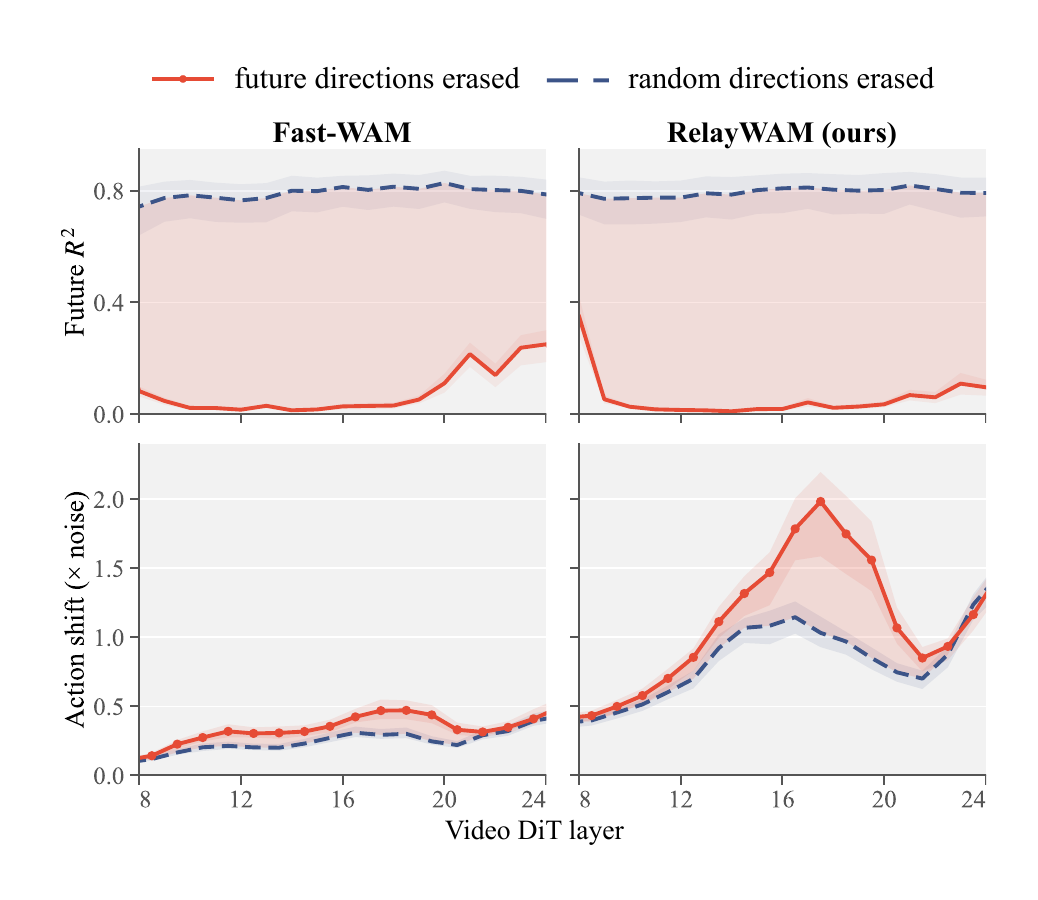}
  \caption{\textbf{Causal erasure of future information in Fast-WAM vs.\ RelayWAM.}
  We erase residual future directions from a sliding window of 4 consecutive Video DiT layers read by the action pathway, leaving the video backbone unchanged.
  \textbf{Top:} held-out $R^2$ of the residual future decoded after future erasure (solid red) versus variance-matched random erasure of the same rank (dashed blue).
  \textbf{Bottom:} resulting action shift normalized by noise-resampling variance, plotted at the center of each erased window.
  While Fast-WAM (left) is insensitive to future erasure, RelayWAM (right) shows a pronounced action shift across intermediate-to-deep layers (layers 12--20), confirming active utilization of predictive cues.
  Shaded bands indicate 95\% confidence intervals over episodes.}
  \label{fig:erasure}
\end{figure}

These findings motivate us to build RelayWAM, which introduces an explicit interface to bridge the two pathways. It introduces a transition space between the intermediate-to-deep layers of the Video DiT and the Action DiT.
Since the future information required for action generation varies across action steps and execution states, we condition transition tokens on the task instruction, robot state, and current action estimate.
This allows the tokens to dynamically extract the future information most relevant to each action step from the future-supervised visual features, relay it across depth, and deliver it to the action pathway.

With the transition space, the same future supervision becomes far more effective.
In a controlled $2{\times}2$ study, future supervision improves LIBERO-Plus success by 5.2 points on the base architecture, but by 15.6 points with the transition space.
Fast-WAM~\cite{fastwam} likewise generates no future at inference; RelayWAM raises its LIBERO-Plus success rate from 51.50\% to 72.03\%.
This also surpasses OpenWAM-$\alpha$ (69.2\%)~\cite{openwam}, which jointly generates future and action at inference.

\noindent\textbf{Contributions.} Our contributions are as follows:
\begin{itemize}
\item We characterize how future supervision shapes shared representations and how the resulting information relates to action generation, providing a representation-level analysis of future-supervised WAMs.
\item We introduce training-focus attribution, together with linear probes and causal erasure, to investigate the writing and reading behaviors of future-supervised representations.
\item We build RelayWAM, which adds an explicit interface that converts future information into a form usable for action. It makes future supervision substantially more effective and achieves strong generalization in simulation and on real robots.
\end{itemize}

%% file: sec/2_relatedwork.tex
\section{Related Work}
\label{sec:related}

\paragraph{Vision-language-action policies.}
VLA policies~\cite{rt2,openvla,pi0,pi05,pi07,rdt,gr00t,smolvla,geminirobotics,dexvla,cotvla,spatialvla,univla,spatialforcing,vlastem,coalitionskill,dynahmrc} generate actions from the current language instruction and visual observation.
They are supervised mainly by robot trajectories, whose scale and diversity are far below internet-scale data, and policies trained only to predict actions tend to rely on the current visual appearance rather than on how the environment will evolve under interaction.
WAMs address this by using pretrained video models as world models and adding future-prediction objectives that capture object dynamics and state transitions, which improves action generation and cross-task generalization~\cite{unipi,wan,robodreamer,dreamgen}.

\paragraph{World action models.}
WAMs differ in how future information enters action generation.
Explicit WAMs first generate future observations and then decode actions through an inverse dynamics model or an action head~\cite{unipi,gen2act,vidar,videopolicy,mimicvideo,gr2,genieenvisioner,dreamzero}.
Implicit WAMs avoid pixel-level generation by compressing the future into latent representations or transition tokens, exploited through future prediction or joint future--action modeling~\cite{motus,lawam,cosmospolicy,videovla,vpp,uwm,uva,worldvla,rynnvla002,dreamvla,gigaworldpolicy,nativevap,trajwam,wam4d,geosemwam,geo4dwam,pointworld}.
A third line keeps future prediction only as a training objective on the shared backbone and generates actions from the current observation alone~\cite{flare,fastwam,vlajepa,lightwam,jepawam}; we refer to these as \emph{future-free inference} (FFI) WAMs.
FFI WAMs keep the benefits of future prediction at no extra deployment cost, but implicitly assume that the action pathway uses the predictive representations in the shared backbone.
RelayWAM shows that the action reads this information only weakly and introduces transition layers that convert future-supervised representations into a form usable for action.

%% file: sec/3_Analysis.tex
\section{Analysis}
\label{sec:analysis}

While attention weights show where tokens look, they do not measure how much each visual feature actually contributes to a training objective.
To capture this objective-dependent reliance, we introduce \emph{training-focus attribution}, which quantifies the contribution of visual features to each objective through their activations and gradients.

We analyze two models: \textbf{A1}, Fast-WAM jointly trained with action and future objectives, $\mathcal{L}_{\text{action}}+\lambda_{\text{future}}\mathcal{L}_{\text{future}}$, and \textbf{A0}, an otherwise identical model trained with the action objective alone, $\mathcal{L}_{\text{action}}$.
Details of all analyses are provided in the supplementary material.

\subsection{What future supervision writes}
\label{sec:analysis_write}

We first measure how much future information the visual features hold.
A linear probe~\cite{alain2017probe,hewitt2019control} decodes the future latent change $\Delta z$ from the pooled visual features of layer $l$.
The contribution of future supervision is the gap in held-out $R^2$ between the two models:
\begin{equation}
  \Delta R^2_l = R^2_{l,\text{A1}} - R^2_{l,\text{A0}}.
  \label{eq:probe_gap}
\end{equation}
The gap is small in shallow layers and widens with depth (Fig.~\ref{fig:teaser}, top right).
Decodability alone does not reveal which visual content each objective relies on.
We therefore attribute each objective to the visual values $V^v_l$ read by the action attention at layer $l$~\cite{shrikumar2017deeplift,sundararajan2017ig,gradcam}:
\begin{equation}
  s^{\mathcal{L}}_l(p) = \sum_{c} \Bigl|\, V^v_{l,p,c}\, \frac{\partial \mathcal{L}}{\partial V^v_{l,p,c}} \Bigr|,
  \label{eq:attribution}
\end{equation}
where $p$ indexes tokens, $c$ indexes channels, and $\mathcal{L}\in\{\mathcal{L}_{\text{action}},\mathcal{L}_{\text{future}}\}$.
This score bounds the first-order change in $\mathcal{L}$ when token $p$ is removed~\cite{molchanov2017pruning}.
In shallow layers, the two maps coincide.
In deep layers, the action focus contracts onto the gripper and the manipulated object, while the future focus remains spread across the entire scene, background included (Fig.~\ref{fig:gradplane}, bottom).
Token attribution compares spatial reliance, but representations also encode information along feature directions.
At each layer, we extract the top-$k$ principal directions of the action and future gradients, $U^A_l$ and $U^F_l$, and measure the share of held-out gradient energy falling into each.
The two gradients share directions in shallow layers but concentrate on nearly disjoint directions in deep layers~\cite{pcgrad} (Fig.~\ref{fig:gradplane}, top left).

\subsection{What the action pathway reads}
\label{sec:analysis_read}

\begin{figure*}[t]
  \centering
  \includegraphics[width=\textwidth]{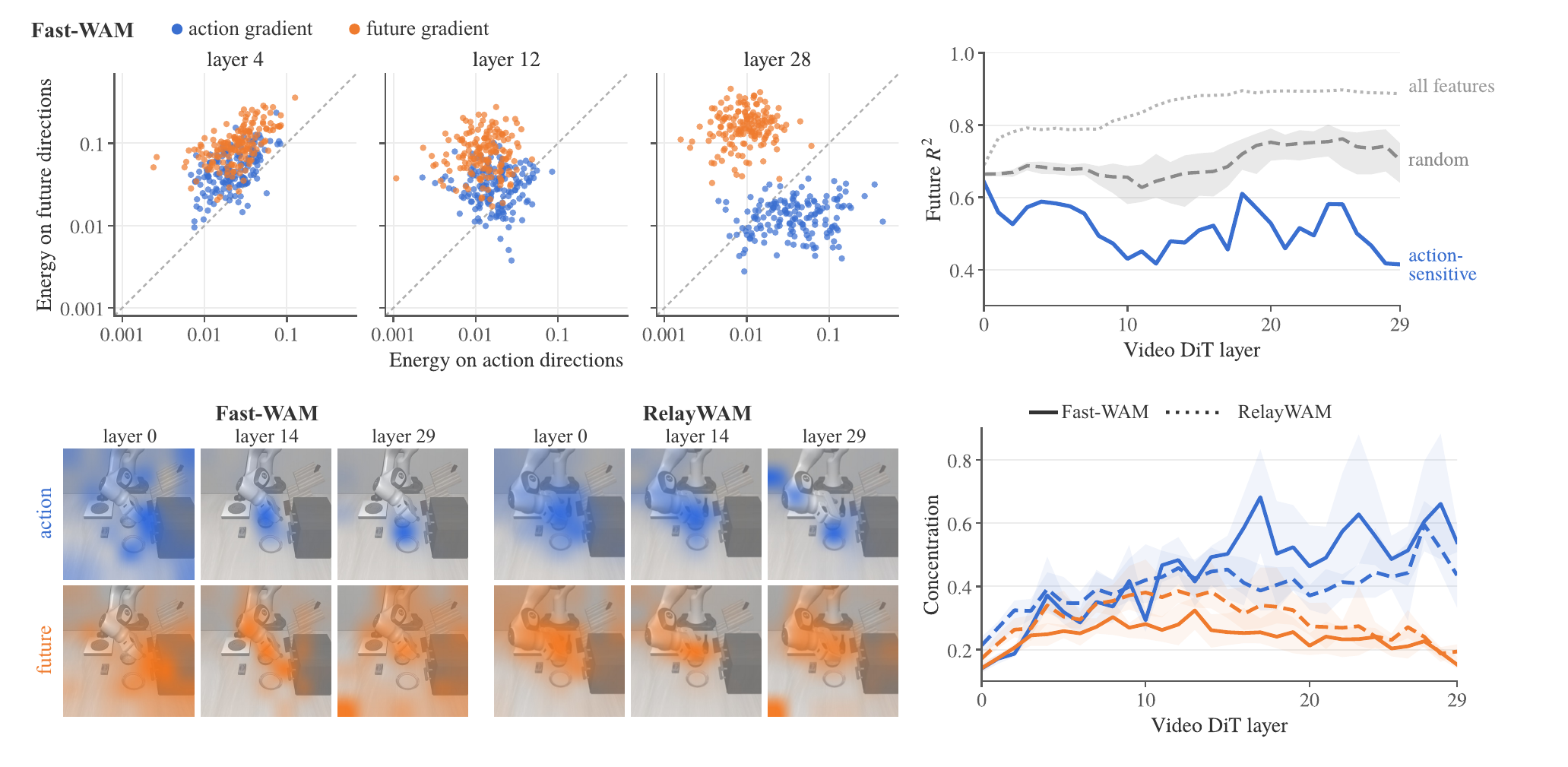}
  \caption{\textbf{Action and future supervision diverge across depth in Fast-WAM.}
  \textbf{Top left:} gradient energy projected onto top-$k$ action and future directions shows that the two objectives share subspaces in shallow layers (layer 4) but become nearly orthogonal in deep layers (layer 28).
  \textbf{Top right:} held-out future $R^2$ decoded from action-sensitive subspaces drops well below random subspaces of equal dimension as depth increases.
  \textbf{Bottom:} training-focus maps on ``open the middle drawer'' and top-10\% focus concentration averaged over five tasks. Unlike Fast-WAM (solid), RelayWAM (dashed) keeps both objectives focused on task-relevant regions.}
  \label{fig:gradplane}
\end{figure*}

We refer to the columns of $U^A_l$ as the \emph{action-sensitive directions}~\cite{kaufman2014null,elsayed2016reorganization}, along which perturbations to the representation induce the largest first-order change in $\mathcal{L}_{\text{action}}$.
Decoding $\Delta z$ from these directions alone and comparing against variance-matched random subspaces of equal dimension reveals that both decode the future equally well in shallow layers; with depth, however, action-sensitive subspaces carry substantially less future information than random ones (Fig.~\ref{fig:gradplane}, top right).
We next test whether action generation causally depends on this predictive information by erasing the \emph{residual future}—the component of $\Delta z$ not explained by the ground-truth action and proprioceptive state—from the features read by the action attention~\cite{ravfogel2020null,leace}.
In Fast-WAM, erasing these directions removes the residual future from the features but shifts the action less than resampling the denoising noise does—in stark contrast to RelayWAM, where the same erasure induces a pronounced action shift across layers 12--20 (Fig.~\ref{fig:erasure}).

Beyond Fast-WAM, we further apply the same analysis protocol to two additional FFI WAMs, JEPA-WAM~\cite{jepawam} and CSWAM~\cite{liu2026cswam}, and observe a consistent pattern: the training focus of the two objectives diverges with depth (Fig.~\ref{fig:teaser}, left), and future information decoded from action-sensitive directions falls below that of random subspaces.

\begin{center}
\fbox{\parbox{0.94\linewidth}{%
\textbf{Finding.} In FFI WAMs, future supervision writes rich predictive information into deep visual features, but along directions the action pathway barely reads.
We term this the \emph{representation--utilization gap}.
The bottleneck lies not in the predictive capacity of the backbone, but in the missing conversion of future information into a form usable for action.}}
\end{center}

%% file: sec/4_method.tex
\section{RelayWAM}
\label{sec:method}

\subsection{Overview}
\label{sec:RelayWAM_overview}

\begin{figure*}[t]
    \centering
    \includegraphics[width=\textwidth]{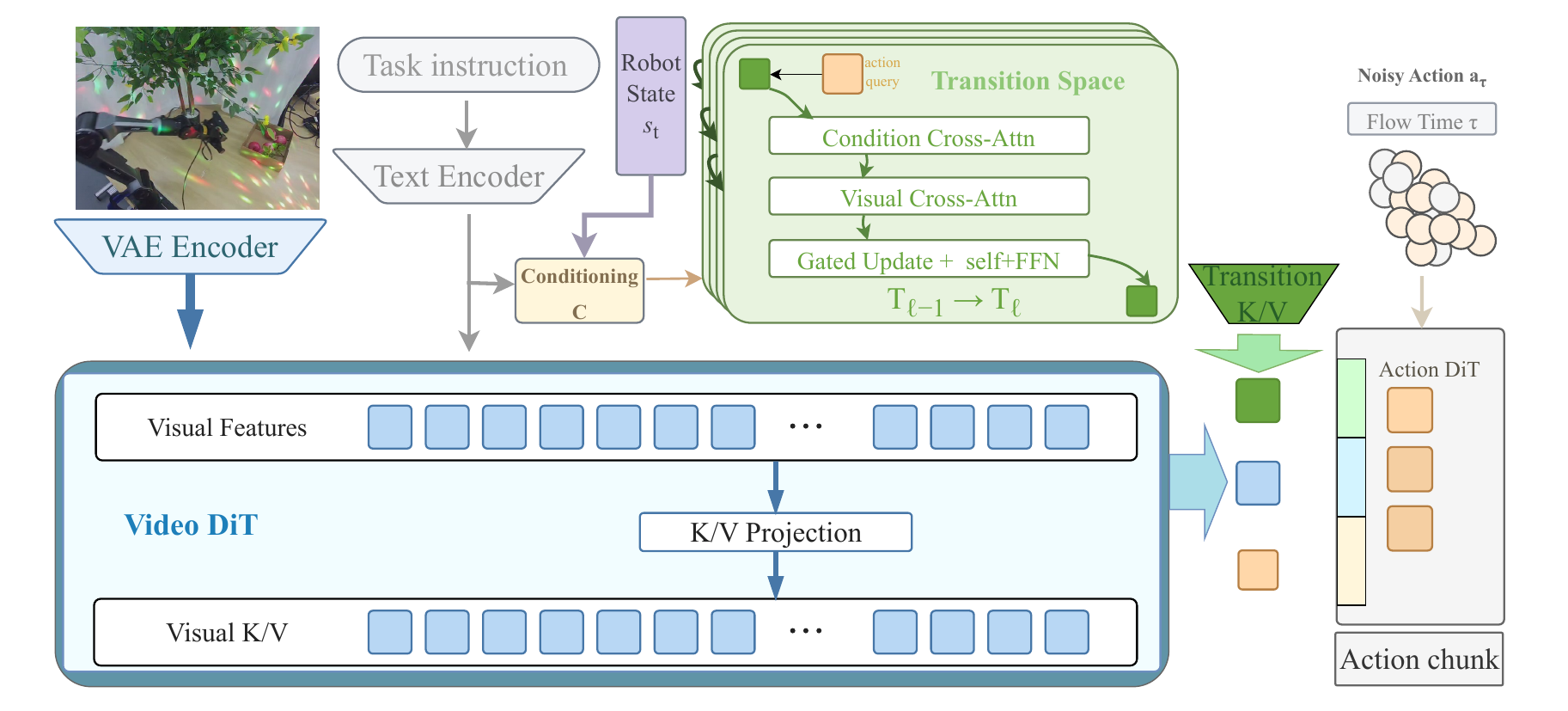}
    \caption{
        \textbf{Architecture of RelayWAM.}
        Attached to selected intermediate and deep layers of the Video DiT, recurrent transition layers extract step-relevant future information from visual representations conditioned on the task instruction, robot state, current action estimate, and flow time.
        The extracted cues are progressively accumulated across depth into action-aligned transition tokens $\mathbf{T}_{l}$, projected into transition keys and values $(\hat{\mathbf{K}}_{l},\hat{\mathbf{V}}_{l})$, and injected directly into the native Action DiT attention.
        This enables action generation to leverage multi-depth predictive representations without generating future observations at inference time.
    }
    \label{fig:RelayWAM_overview}
\end{figure*}

To bridge the representation--utilization gap identified in Sec.~\ref{sec:analysis}, RelayWAM introduces an explicit interface that extracts step-relevant future cues from the visual backbone and converts them into action-usable representations (Fig.~\ref{fig:RelayWAM_overview}).
Specifically, transition layers attached to intermediate and deep layers of the Video DiT extract predictive visual cues conditioned on the task instruction and robot state, accumulate them across depth into transition tokens aligned with the action steps, and inject these tokens directly into the native action attention, while keeping future frames purely as a training signal.

\paragraph{Backbone.}
RelayWAM builds on an FFI WAM~\cite{fastwam} in which a pretrained Video DiT~\cite{dit,wan} and an Action DiT, each with $L$ layers ($L=30$), are coupled as a Mixture-of-Transformers (MoT)~\cite{mot}.
Given the current observation $o_t$, the language instruction $c$, and the proprioceptive state $s_t\in\mathbb{R}^{D_s}$, the Action DiT generates an action chunk of $H$ steps via flow matching~\cite{flowmatching,rectifiedflow} with flow time $\tau\in[0,1]$.
At layer $l$, let $\mathbf{X}^{v}_{l}\in\mathbb{R}^{N\times d_v}$ denote the $N$ current-frame video tokens and $\mathbf{X}^{a}_{l}\in\mathbb{R}^{H\times d_a}$ the action tokens, one per action step.
Video tokens attend only to themselves, so the Video DiT remains independent of the action pathway.
Action tokens receive visual features solely through the action attention~\cite{vaswani2017attention}:
\begin{equation}
    \operatorname{softmax}
    \left(
        \frac{
            \mathbf{Q}^{a}_{l}
            [\mathbf{K}^{v}_{l};\mathbf{K}^{a}_{l}]^{\top}
        }{
            \sqrt{d_h}
        }
    \right)
    [\mathbf{V}^{v}_{l};\mathbf{V}^{a}_{l}],
    \label{eq:native_attention}
\end{equation}
where $\mathbf{Q}^{a}_{l}$, $\mathbf{K}^{a}_{l}$, and $\mathbf{V}^{a}_{l}$ are the action queries, keys, and values, $(\mathbf{K}^{v}_{l},\mathbf{V}^{v}_{l})$ are the keys and values of the current-frame video tokens, $[\,\cdot\,;\cdot\,]$ denotes concatenation along the token axis, and $d_h$ is the head dimension.

\subsection{Transition layers}
\label{sec:transition_modeling}

RelayWAM attaches a transition layer to each layer $l \in \{l_0,\dots,L-1\}$ ($l_0=14$)~\cite{layerroute}, where future-predictive representations become prominent (Sec.~\ref{sec:analysis_write}).
These layers maintain a set of transition tokens $\mathbf{T}_{l}\in\mathbb{R}^{H\times d_a}$, one per action step.
Starting from $\mathbf{T}_{l_0-1}=\mathbf{0}$, the transition layer at depth $l$ updates $\mathbf{T}_{l-1}$ to $\mathbf{T}_{l}$ using the action tokens $\mathbf{X}^{a}_{l}$ and the current-frame video tokens $\mathbf{X}^{v}_{l}$.
Below, $\operatorname{CrossAttn}(\mathbf{Q},\mathbf{Y})$ denotes multi-head cross-attention with queries projected from $\mathbf{Q}$ and keys and values projected from $\mathbf{Y}$~\cite{perceiverio}.

\paragraph{Task- and state-conditioned extraction.}
We first encode the task instruction and robot proprioceptive state into conditioning tokens $\mathbf{C}=[\mathbf{C}^{\mathrm{lang}};\mathbf{C}^{\mathrm{state}}]$.
The language tokens $\mathbf{C}^{\mathrm{lang}}$ are obtained via a linear projection of the instruction embedding from the backbone's text encoder~\cite{t5}.
For the state, each of the $D_s$ proprioceptive dimensions is projected by a lightweight MLP and added to a learned joint-index embedding, allowing $\mathbf{C}^{\mathrm{state}}$ to represent individual joints.
$\mathbf{C}$ is computed once per control step and shared across all transition layers.
The extraction query encodes the action step being denoised:
\begin{equation}
    \mathbf{Q}_{l}
    =
    \phi(\mathbf{X}^{a}_{l})
    +
    \psi(\mathbf{T}_{l-1})
    +
    \mathbf{E}_{l}(\tau),
    \label{eq:transition_query}
\end{equation}
where $\phi$ and $\psi$ are linear projections applied after layer normalization, and $\mathbf{E}_{l}(\tau)$ sums the embeddings of the action-step index, layer index $l$, and flow time $\tau$.
The query first attends to the task and state context,
\begin{equation}
    \tilde{\mathbf{Q}}_{l}
    =
    \mathbf{Q}_{l}
    +
    \operatorname{CrossAttn}
    \bigl(
        \mathbf{Q}_{l},
        \mathbf{C}
    \bigr),
    \label{eq:transition_condition}
\end{equation}
and then reads out visual cues from the current-frame video tokens at the same depth:
\begin{equation}
    \mathbf{R}_{l}
    =
    \operatorname{CrossAttn}
    \bigl(
        \tilde{\mathbf{Q}}_{l},
        \operatorname{LN}_{l}(\mathbf{X}^{v}_{l})
    \bigr),
    \label{eq:transition_readout}
\end{equation}
where $\operatorname{LN}_{l}$ is a layer-specific layer normalization and the video keys incorporate 2D sinusoidal position encodings~\cite{vaswani2017attention,c2rope,sope}.
Thus, the task, robot state, and action step jointly determine \emph{what} to query, while the retrieved content $\mathbf{R}_{l}$ is drawn directly from the future-supervised visual features rather than the action pathway.

\begin{table*}[t]
    \centering
    \small
    \setlength{\tabcolsep}{5.0pt}
    \renewcommand{\arraystretch}{1.15}

    \begin{tabular*}{\textwidth}{@{\extracolsep{\fill}}clcccccccc@{}}
        \toprule
        \multirow{2}{*}{\textbf{Paradigm}} & \multirow{2}{*}{\textbf{Method}}
        & \multicolumn{8}{c}{\textbf{LIBERO-Plus Success Rate (\%)}} \\
        \cmidrule(l){3-10}
        & & Bg.\,$\uparrow$ & Cam.\,$\uparrow$ & Lang.\,$\uparrow$ & Light\,$\uparrow$
          & Layout\,$\uparrow$ & Robot\,$\uparrow$ & Noise\,$\uparrow$ & Total\,$\uparrow$ \\
        \midrule

        \multirow{3}{*}{VLA}
        & OpenVLA~\cite{openvla}         & 34.8 & 0.8  & 23.0 & 8.1  & 28.5 & 3.5  & 15.2 & 15.6 \\
        & $\pi_{0}$~\cite{pi0}           & 81.4 & 13.8 & 58.8 & 85.0 & 68.9 & 6.0  & \snd{79.0} & 53.6 \\
        & OpenVLA-OFT~\cite{openvlaoft}  & \fst{93.3} & \fst{56.4} & 79.5 & 88.7 & 74.2 & 31.9 & 75.8 & \snd{69.6} \\
        \midrule

        \multirow{2}{*}{Joint}
        & Fast-WAM-Joint~\cite{fastwam}    & 44.9 & 34.0 & \fst{88.9} & 90.0 & 73.4 & 55.1 & 33.3 & 59.1 \\
        & OpenWAM-$\alpha$~\cite{openwam}  & \snd{87.1} & 33.8 & \snd{88.0} & \fst{97.0} & \snd{77.5} & \fst{76.1} & 39.8 & 69.2 \\
        \midrule

        \multirow{3}{*}{FFI}
        & Fast-WAM~\cite{fastwam}    & 53.7 & 16.4 & 68.9 & 78.2 & 60.7 & 44.5 & 37.7 & 51.5 \\
        & DreamWAM~\cite{dreamwam}   & 64.7 & 30.4 & 77.3 & 85.9 & 70.0 & 59.2 & 56.7 & 63.4 \\
        \cmidrule(l){2-10}
        & \textbf{RelayWAM (Ours)}   & 54.7 & \snd{47.0} & 87.1 & \snd{93.2} & \fst{80.7} & \snd{61.6} & \fst{81.1} & \fst{72.0} \\
        \bottomrule
    \end{tabular*}

    \caption{
        \textbf{Results on LIBERO-Plus~\cite{liberoplus} under distribution shifts}~--
        We report success rates (\%) across seven perturbation types; Total is the
        success rate over all evaluation episodes, following~\cite{liberoplus}.
        \textit{Joint} denotes joint future--action decoding at inference,
        while \textit{FFI} denotes future-free inference without explicit
        future generation at test time. Darker and lighter blue cells mark the
        best and second-best result in each column.
    }
    \label{tab:libero_plus}
\end{table*}

\paragraph{Relay update.}
The extracted content is gated by flow time and integrated into the incoming transition tokens via a transition block:
\begin{equation}
    \mathbf{T}_{l}
    =
    \operatorname{TransBlock}_{l}
    \bigl(
        \mathbf{T}_{l-1}
        +
        g_{l}(\tau)\,\mathbf{R}_{l}
    \bigr),
    \label{eq:transition_update}
\end{equation}
where $g_{l}(\tau)\in(0,1)$ is a scalar sigmoid gate derived from the flow-time embedding with a layer-specific bias, and $\operatorname{TransBlock}_{l}$ is a pre-norm Transformer block consisting of self-attention over the $H$ transition tokens (with rotary position encodings along the action-step axis) and a feed-forward network.
By accumulating extracted cues from layer $l_0$ onward, $\mathbf{T}_{l}$ aggregates multi-depth predictive representations without redundantly re-reading earlier layers.
To keep the interface lightweight (adding only 34.96M parameters, $<0.6\%$ of the 6.02B backbone), all transition layers share their projection, attention, and feed-forward weights, retaining only lightweight layer-specific normalization, scaling, and bias vectors so that the interface operates recurrently across depth.

The updated tokens are then read directly by the action attention at the same depth.
We project $\operatorname{LN}(\mathbf{T}_{l})$ into transition keys $\hat{\mathbf{K}}_{l}$ and values $\hat{\mathbf{V}}_{l}$, applying RMS normalization and step-aligned rotary position encodings to $\hat{\mathbf{K}}_{l}$ so that the $h$-th transition token aligns with the $h$-th action token.
The native action attention in Eq.~\eqref{eq:native_attention} is thus extended to
\begin{equation}
    \operatorname{softmax}
    \left(
        \frac{
            \mathbf{Q}^{a}_{l}
            [\mathbf{K}^{v}_{l};\hat{\mathbf{K}}_{l};\mathbf{K}^{a}_{l}]^{\top}
        }{
            \sqrt{d_h}
        }
        +
        \mathbf{B}_{l}
    \right)
    [\mathbf{V}^{v}_{l};\hat{\mathbf{V}}_{l};\mathbf{V}^{a}_{l}],
    \label{eq:transition_attention}
\end{equation}
where $\mathbf{B}_{l}$ adds a learned per-head routing bias to the logits of the transition keys and is zero elsewhere.
Consequently, action generation integrates converted future information directly within its native attention mechanism without requiring any external decoder, allowing each attention head to dynamically balance raw visual features against the relayed predictive tokens.

\subsection{Training and inference}
\label{sec:training_inference}

RelayWAM retains the exact training objective $\mathcal{L}_{\text{action}}+\lambda_{\text{future}}\mathcal{L}_{\text{future}}$ of FFI WAMs without introducing any auxiliary loss on the transition layers: the feature conversion is learned end-to-end from $\mathcal{L}_{\text{action}}$, whose gradients also propagate back into $\mathbf{X}^{v}_{l}$ through the cross-attention readout, enabling both objectives to jointly shape the visual backbone.
At inference, RelayWAM generates no future frames.
The Video DiT runs only once on the current observation, and its visual keys/values and the conditioning tokens $\mathbf{C}$ are cached across all denoising steps.
Since the transition tokens depend on the current action estimate and flow time, only the lightweight transition layers are rerun at each denoising step, recomputing $\mathbf{T}_{l}$ from $\mathbf{T}_{l_0-1}=\mathbf{0}$ and the resulting transition keys/values $(\hat{\mathbf{K}}_{l},\hat{\mathbf{V}}_{l})$.

%% file: sec/5_experiments.tex
\section{Experiments}
\label{sec:experiments}

\begin{figure*}[t]
  \centering
  \includegraphics[height=2.25in]{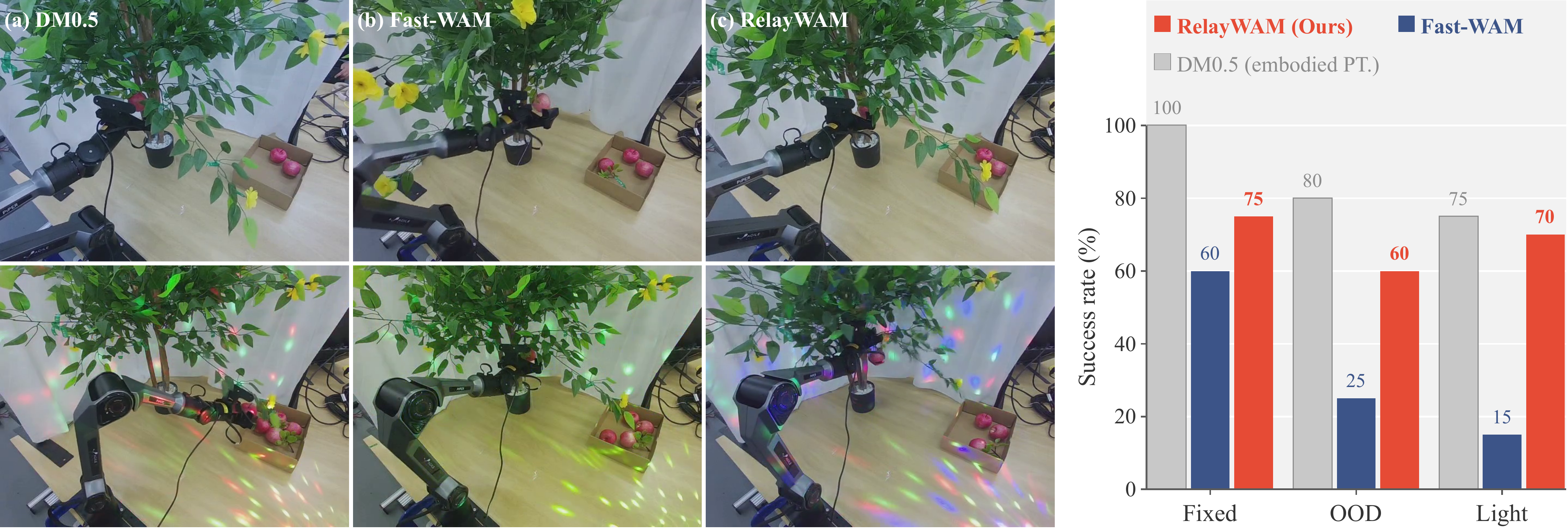}
  \caption{\textbf{Real-world apple picking on a Piper robotic arm.}
  \textbf{Left:} qualitative rollouts under out-of-distribution apple placements with leaf occlusion (top) and colored ambient lighting perturbations (bottom).
  \textbf{Right:} success rates (\%) across 20 trials per setting. Without any large-scale embodied pretraining (unlike DM0.5), RelayWAM outperforms Fast-WAM across all settings and maintains strong robustness under severe illumination shifts (70\% vs.\ 15\%).}
  \label{fig:real_setup}
\end{figure*}

\subsection{Experimental setup}
\label{sec:setup}

\paragraph{Implementation details.}
RelayWAM builds upon a MoT-style~\cite{mot} coupled Video DiT--Action DiT backbone~\cite{wan,fastwam} with 6.02B parameters, while the proposed transition space introduces only 34.96M additional parameters ($<0.6\%$ overhead).
To ensure a fair comparison, all our models, including the controlled variants in the ablation study, share the exact same training configuration: 20,000 optimization steps on $8\times$ NVIDIA A100-80GB GPUs with the AdamW optimizer.
At inference, RelayWAM follows the FFI paradigm and directly outputs actions without explicitly generating future observations.

\paragraph{Benchmarks and evaluation.}
We train a single model jointly on the four standard LIBERO~\cite{libero} suites (Spatial, Object, Goal, and Long) and evaluate it across LIBERO, LIBERO-Plus~\cite{liberoplus}, and LIBERO-PRO~\cite{liberopro} using their official rollout protocols.
While the standard suite measures in-distribution manipulation performance, the latter two benchmarks evaluate zero-shot generalization under distribution shifts without additional training: LIBERO-Plus spans seven perturbation types (background, camera viewpoint, language instruction, lighting, object layout, robot initial state, and sensor noise), and LIBERO-PRO covers five generalization dimensions (environment, language, object, position, and task).
We report task success rates (\%) across all evaluations.

\paragraph{Real-world evaluation.}
We further evaluate RelayWAM in the physical world by building an indoor artificial apple tree to emulate an orchard environment, as shown in Fig.~\ref{fig:real_setup}.
In real orchards, variations in fruit and branch positions, partial leaf occlusion, and changing ambient illumination pose significant challenges to policy robustness.
We collect 80 demonstrations, train all compared models on them for 10 epochs, and deploy the resulting policies on a Piper robotic arm.
We evaluate 20 trials under each of three settings: in the \emph{Fixed} setting, the apple remains at the location used during data collection; in the \emph{OOD} setting, the apple is placed at randomized, unseen positions, including locations partially occluded by leaves; and in the \emph{Light} setting, colored ambient lights are added on top of random placements to induce severe illumination shifts.

\begin{table}[t]
    \centering
    \scriptsize
    \setlength{\tabcolsep}{3pt}
    \renewcommand{\arraystretch}{1.10}

    \caption{
        \textbf{Results on LIBERO~\cite{libero}.}
        \textit{VLA} denotes vision-language-action policies without future prediction, \textit{Joint} denotes joint future--action inference, and \textit{FFI} denotes future-free inference.
    }
    \label{tab:libero_results}

    \resizebox{\columnwidth}{!}{%
    \begin{tabular}{@{}llccccc@{}}
        \toprule
        Method & Paradigm & Spatial & Object & Goal & Long & Avg. \\
        \midrule

        $\pi_{0.5}$~\cite{pi05}
        & VLA & 98.8 & 98.2 & 98.0 & 92.4 & 96.9 \\

        \midrule

        LingBot-VA~\cite{lingbotva}
        & Joint & 98.5 & 99.6 & 97.2 & 98.5 & 98.5 \\

        Motus~\cite{motus}
        & Joint & 96.8 & 99.8 & 96.6 & 97.6 & 97.7 \\

        Fast-WAM-Joint~\cite{fastwam}
        & Joint & 99.6 & 99.4 & 98.2 & 96.8 & 98.5 \\

        OpenWAM-$\alpha$~\cite{openwam}
        & Joint & 99.6 & 99.6 & 99.8 & 98.2 & 99.3 \\

        \midrule

        Light-WAM~\cite{lightwam}
        & FFI & 98.2 & 99.6 & 97.8 & 93.0 & 97.2 \\

        Fast-WAM~\cite{fastwam}
        & FFI & 98.2 & 100.0 & 97.0 & 95.2 & 97.6 \\

        \textbf{RelayWAM (Ours)}
        & \textbf{FFI}
        & \textbf{98.4}
        & \textbf{100.0}
        & \textbf{97.6}
        & \textbf{96.8}
        & \textbf{98.2} \\

        \bottomrule
    \end{tabular}%
    }
\end{table}

\begin{table}[t]
    \centering
    \footnotesize
    \setlength{\tabcolsep}{2.5pt}
    \renewcommand{\arraystretch}{1.15}

    \begin{tabular*}{\columnwidth}{@{\extracolsep{\fill}}clcccccc}
        \toprule
        \multirow{2}{*}{\textbf{Paradigm}} & \multirow{2}{*}{\textbf{Method}}
        & \multicolumn{6}{c}{\textbf{LIBERO-PRO Success Rate (\%)\,$\uparrow$}} \\
        \cmidrule(l){3-8}
        & & Env. & Lang. & Obj. & Pos. & Task & Avg. \\
        \midrule

        \multirow{3}{*}{VLA}
        & OpenVLA~\cite{openvla}    & 68.0 & 97.3 & 93.0 & 0.0  & 0.0  & 51.7 \\
        & $\pi_{0}$~\cite{pi0}      & 38.8 & 90.5 & 90.5 & 0.0  & 0.0  & 44.0 \\
        & $\pi_{0.5}$~\cite{pi05}   & 52.8 & 95.8 & 96.0 & 20.8 & 0.8  & 53.2 \\
        \midrule

        \multirow{2}{*}{FFI}
        & Fast-WAM~\cite{fastwam}   & 32.1 & 95.1 & 73.8 & 0.4  & 2.3  & 40.7 \\
        \cmidrule(l){2-8}
        & \textbf{RelayWAM (Ours)}  & 34.3 & 93.9 & 75.2 & 1.4  & 18.8 & 44.7 \\
        \bottomrule
    \end{tabular*}

    \caption{
        \textbf{Results on LIBERO-PRO~\cite{liberopro} under challenging distribution shifts}~--
        We report success rates (\%) across five generalization dimensions. \textit{FFI} denotes future-free inference without explicit future generation at test time.
    }
    \label{tab:libero_pro}
\end{table}

\subsection{Main results}
\label{sec:main_results}

\Cref{tab:libero_results} summarizes the evaluation results on the standard LIBERO benchmark.
RelayWAM achieves an average success rate of \textbf{$98.2\%$}, outperforming existing FFI baselines such as Fast-WAM and Light-WAM while confirming that integrating the lightweight transition space preserves strong in-distribution manipulation accuracy.

While standard in-distribution performance is already close to saturation, RelayWAM's core advantage becomes most pronounced under distribution shifts.
As shown in \Cref{tab:libero_plus}, RelayWAM achieves an overall success rate of \textbf{$72.03\%$} on LIBERO-Plus, outperforming the FFI baseline Fast-WAM by more than $20$ percentage points and delivering substantial gains across camera viewpoint, lighting, object layout, and sensor noise perturbations.
Notably, despite skipping future generation entirely at inference, RelayWAM also surpasses OpenWAM-$\alpha$, which relies on costly test-time future decoding.
This confirms that strong out-of-distribution robustness does not require generating future frames during deployment, provided that predictive representations learned during training are effectively relayed to the action pathway.

A consistent trend is observed on the more challenging LIBERO-PRO benchmark (\Cref{tab:libero_pro}).
RelayWAM improves over Fast-WAM ($44.7\%$ vs.\ $40.7\%$), with particularly pronounced gains in cross-task generalization where baseline FFI models struggle to transfer learned behaviors.

\begin{table}[t]
    \centering
    \small
    \setlength{\tabcolsep}{2pt}
    \renewcommand{\arraystretch}{1.15}

    \begin{tabularx}{\columnwidth}{
        @{}>{\raggedright\arraybackslash}X
        >{\centering\arraybackslash}c
        >{\centering\arraybackslash}c
        >{\centering\arraybackslash}c@{}
    }
        \toprule
        \multirow{2}{*}{\textbf{Variant}}
        & \multicolumn{3}{c}{\textbf{Success Rate (\%)\,$\uparrow$}} \\
        \cmidrule(l){2-4}
        & LIBERO & LIBERO-Plus & LIBERO-PRO \\
        \midrule
        A0 (Base)               & 81.9       & 46.3       & 37.0       \\
        A1 (+ Future Sup.)      & \snd{97.6} & 51.5       & \snd{40.7} \\
        B0 (+ Transition Space) & 96.3       & \snd{56.4} & 39.8       \\
        \midrule
        \textbf{B1 (RelayWAM)}  & \fst{98.2} & \fst{72.0} & \fst{44.7} \\
        \bottomrule
    \end{tabularx}

    \caption{
        \textbf{Ablation study}~--
        Success rates (\%) on LIBERO, LIBERO-Plus, and LIBERO-PRO,
        reported with the same metric as
        Tables~\ref{tab:libero_plus} to~\ref{tab:libero_pro}.
        Darker and lighter blue cells mark the best and
        second-best result in each column.
    }
    \label{tab:ablation}
\end{table}

\subsection{Ablation study}
\label{sec:ablation}

To disentangle the contributions of future supervision and the transition space, we construct a $2\times2$ controlled comparison:
A0 is the Base architecture trained with action supervision alone, A1 adds future supervision, B0 adds the transition space without future supervision, and B1 is the full RelayWAM.
All variants share the same backbone, action objective, training data, and optimization configuration.

\begin{figure}[t]
    \centering
    \includegraphics[width=\columnwidth]{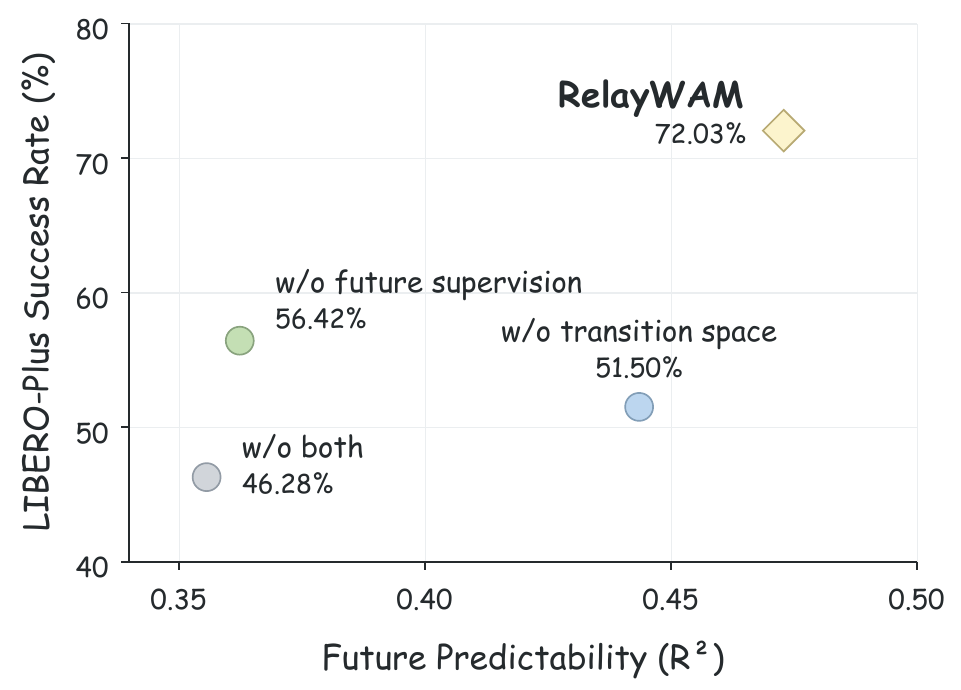}
    \caption{\textbf{Controlled $2{\times}2$ ablation of future supervision and the transition space.}
    LIBERO-Plus success rate (\%) versus layer-20 future predictability ($R^2$) across all four controlled variants. Adding future supervision alone (w/o transition space) raises $R^2$ but yields only a $+5.2$-point success gain over the base model (w/o both), whereas introducing the transition space unlocks the utility of predictive features and amplifies the gain to $+15.6$ points ($72.03\%$ overall).}
    \label{fig:ablation}
\end{figure}

As shown in \Cref{tab:ablation} and \Cref{fig:ablation}, the transition space amplifies the benefit of future supervision.
On LIBERO-Plus, future supervision brings only a modest gain on the Base architecture ($+5.2$ points), but roughly triples this gain when paired with the transition space ($+15.6$ points).
LIBERO-PRO shows the same pattern, while in-distribution LIBERO scores are near saturation for A1, B0, and B1.

Representation-level probing helps explain this synergy.
Reusing the layer-20 probe from \Cref{sec:analysis_write}, we find that future supervision raises future predictability more with the transition space than without it, suggesting that the cross-attention readout makes the visual features more responsive to future supervision rather than merely preserving them.
Moreover, actions are already highly decodable from the transition tokens at the entrance of the transition space in RelayWAM, whereas in B0 action decodability builds up only gradually with depth.
Together, these results suggest that \emph{future supervision writes predictive cues into the backbone, and the transition space organizes them into action-usable representations early}, which translates into stronger robustness under distribution shifts.

\subsection{Real-world results}
\label{sec:real_results}

We further compare RelayWAM against Fast-WAM~\cite{fastwam} and DM0.5~\cite{dm05}, a strong VLA baseline built upon large-scale embodied pretraining, with all models trained on the exact same 80 real-world demonstrations.
As shown in \Cref{fig:real_setup}, RelayWAM consistently outperforms Fast-WAM across every evaluation setting, and its performance advantage widens progressively as environmental perturbations intensify from the \emph{Fixed} setting to unseen \emph{OOD} placements.
The \emph{Light} setting separates the two FFI models most dramatically: under colored ambient illumination, Fast-WAM's success rate drops sharply to $15\%$, whereas RelayWAM maintains a $70\%$ success rate—nearly matching the embodied-pretrained DM0.5 ($75\%$) without utilizing any large-scale embodied pretraining.
Since RelayWAM and Fast-WAM share the exact same backbone and differ solely by the 34.96M transition layers, this substantial real-world robustness gain directly stems from converting future-predictive features into representations that the action pathway can readily read, echoing the strong lighting robustness observed on LIBERO-Plus.

%% file: sec/6_llm_statement.tex
%
\section*{LLM Usage Statement}
We used Claude (Anthropic) for three purposes.
First, it translated author-written drafts into English and refined grammar and sentence structure.
Second, it assisted in writing and debugging analysis and plotting scripts.
Third, it helped refine the layout and styling of author-designed figures.
All technical content, claims, and final wording were determined and verified by the authors.
All scripts were reviewed, executed, and validated by the authors.
Claude was not used to design the method, design or run experiments, create data, or draw scientific conclusions.
The authors take full responsibility for all content of this paper.

%% file: sec/X_suppl.tex
\clearpage
\setcounter{page}{1}
\maketitlesupplementary

\appendix

\section{Analysis Protocol and Future-State Probing}
\label{sec:supp_setup}
\setcounter{equation}{0}
\renewcommand{\theequation}{A\arabic{equation}}

To systematically dissect how future-predictive representations are written and read across network depth, we establish a unified diagnostic protocol applied throughout Appendices~\ref{sec:supp_attr}--\ref{sec:supp_erase}.

\paragraph{Evaluated models and notation.}
We compare two controlled models sharing the identical backbone, training data, and optimization setup (Sec.~\ref{sec:method}): \textbf{A1} (Fast-WAM trained jointly with $\mathcal{L}_{\text{action}}+\lambda_{\text{future}}\mathcal{L}_{\text{future}}$) and \textbf{A0} (trained with $\mathcal{L}_{\text{action}}$ alone). At each Video DiT layer $l \in \{0,\dots,29\}$, $X^v_l \in \mathbb{R}^{N \times d_v}$ denotes the current-frame visual tokens ($d_v = 3072$), and $V^v_l$ denotes the corresponding visual values attended to by the Action DiT.

\paragraph{Dataset and episode-level cross-validation.}
All representation analyses are conducted on 800 temporal windows sampled from 100 LIBERO episodes at flow time $\tau = 0.5$ (with consistent trends verified on 400-window subsets at $\tau \in \{0.25, 0.75\}$). All probes and subspaces are fitted and evaluated using 5-fold cross-validation partitioned strictly at the episode level, while causal erasers (Appendix~\ref{sec:supp_erase}) are evaluated on an independent set of held-out episodes.

\paragraph{Quantifying written future information.}
To isolate dynamic state transitions from static scene appearance, we define the prediction target as the latent displacement between the current and future frames:
\begin{equation}
  \Delta z = \mathrm{Pool}(z_{t+1}) - \mathrm{Pool}(z_t).
  \label{eq:supp_delta_z}
\end{equation}
Because scene context and robot proprioception partially constrain near-term dynamics even without auxiliary supervision, we isolate the net contribution of future supervision via the decodability gap $\Delta R^2_l = R^2_{l,\text{A1}} - R^2_{l,\text{A0}}$ (Eq.~\eqref{eq:probe_gap}). For each layer $l$, we train a standardized ridge regressor $f_l$ on the spatially pooled visual tokens $\bar{x}_l$, selecting the regularization strength via inner cross-validation over 15 log-spaced values in $[10^{-2}, 10^{5}]$. Held-out performance is measured by the coefficient of determination:
\begin{equation}
  R^2_l = 1 - \frac{\sum_n \lVert \Delta z_n - f_l(\bar{x}_{l,n}) \rVert^2}{\sum_n \lVert \Delta z_n - \overline{\Delta z} \rVert^2}.
  \label{eq:supp_probe}
\end{equation}

\section{Training-Focus Attribution}
\label{sec:supp_attr}
\setcounter{equation}{0}
\renewcommand{\theequation}{B\arabic{equation}}

\paragraph{Formulation.}
Raw attention weights indicate token routing but do not measure sensitivity to the training objective. We therefore quantify objective-specific visual reliance directly on the value representations $V^v_l$ read by the action pathway. Let $V^{v,(p)}_l$ denote $V^v_l$ with token $p$ masked to zero. By first-order Taylor expansion, removing token $p$ perturbs an objective $\mathcal{L} \in \{\mathcal{L}_{\text{action}}, \mathcal{L}_{\text{future}}\}$ by
\begin{equation}
  \mathcal{L}\bigl(V^{v,(p)}_l\bigr) - \mathcal{L}\bigl(V^v_l\bigr) \approx -\sum_{c} V^v_{l,p,c}\, \frac{\partial \mathcal{L}}{\partial V^v_{l,p,c}}.
  \label{eq:supp_taylor}
\end{equation}
Summing channel-wise magnitudes as in Eq.~\eqref{eq:attribution} yields a tight upper bound on the first-order loss perturbation, capturing total token influence without cross-channel sign cancellation.

\paragraph{Computation and spatial concentration.}
Both attribution maps are computed from a single forward pass by back-propagating $\mathcal{L}_{\text{action}}$ and $\mathcal{L}_{\text{future}}$ independently onto the same tensor $V^v_l$, ensuring identical input and network states. To track spatial selectivity across depth, we normalize each map to unit mass and compute the concentration score held by the top 10\% most influential tokens ($\mathcal{T}_{10}$):
\begin{equation}
  C^{\mathcal{L}}_l = \sum_{p\in\mathcal{T}_{10}} \frac{s^{\mathcal{L}}_l(p)}{\sum_{p'} s^{\mathcal{L}}_l(p')},
  \label{eq:supp_concentration}
\end{equation}
where a uniform spatial distribution yields $C^{\mathcal{L}}_l = 0.1$.

\paragraph{Layer-wise divergence.}
As shown in Fig.~\ref{fig:gradplane} (bottom right), both objectives initially exhibit broad, overlapping spatial reliance in shallow layers ($C^{\text{action}}_l \approx C^{\text{future}}_l$). With increasing depth, $C^{\text{action}}_l$ rises steadily as action supervision localizes onto the end-effector and target object, whereas $C^{\text{future}}_l$ remains low and distributed across the background scene.

\section{Subspace Alignment and Inverted Probing}
\label{sec:supp_read}
\setcounter{equation}{0}
\renewcommand{\theequation}{C\arabic{equation}}

\paragraph{Objective gradient subspaces.}
For each window $n$, we compute the token-aggregated gradient with respect to the layer-$l$ visual representation,
\begin{equation}
  g_n = \sum_{p} \frac{\partial \mathcal{L}}{\partial X^v_{l,p}} \in \mathbb{R}^{d_v},
  \label{eq:supp_grad}
\end{equation}
which characterizes the update direction induced by objective $\mathcal{L}$. Applying singular value decomposition to the stacked training gradients yields the top-$k$ principal action-sensitive directions $U^A_l \in \mathbb{R}^{d_v \times k}$ and future-sensitive directions $U^F_l \in \mathbb{R}^{d_v \times k}$ ($k=8$). For any held-out gradient $g$, the fraction of gradient energy captured by an orthonormal basis $U$ is
\begin{equation}
  e(g;U) = \frac{\lVert U^{\top} g \rVert_2^2}{\lVert g \rVert_2^2}.
  \label{eq:supp_energy}
\end{equation}
While an isotropic random $k$-dimensional subspace captures $k/d_v \approx 0.003$ of the gradient energy in expectation, $U^A_l$ and $U^F_l$ each capture up to 45\% of their respective objective's energy in layer 29 ($>100\times$ chance) while remaining nearly orthogonal to one another (Fig.~\ref{fig:gradplane}, top left).

\paragraph{Subspace future decodability.}
To test whether action-sensitive directions carry predictive cues, we project $\bar{x}_l$ onto any $k$-dimensional orthonormal basis $S$ and fit a ridge probe $f_S(S^{\top}\bar{x}_l)$ against the leading principal component of $\Delta z$, denoted $y = u_1^{\top}\Delta z$:
\begin{equation}
  R^2_l(S) = 1 - \frac{\sum_n \bigl( y_n - f_S(S^{\top}\bar{x}_{l,n}) \bigr)^2}{\sum_n ( y_n - \bar{y} )^2}.
  \label{eq:supp_subspace_r2}
\end{equation}
We further construct interpolated subspaces $S_{l,j}$ that smoothly transition from covariance-matched random directions ($j=0$) to purely action-sensitive directions ($j=k$):
\begin{equation}
  S_{l,j} = \mathrm{orth}\bigl([\,U^A_{l,1:j},\ \Sigma_l^{1/2}\Omega_{k-j}\,]\bigr), \quad j=0,\dots,k,
  \label{eq:supp_mixed}
\end{equation}
where $\Sigma_l$ is the empirical feature covariance and $\Omega_{k-j}$ is a standard Gaussian matrix (averaged over 12 draws for $j<k$). Matching the data covariance ensures that baseline subspaces span high-variance feature directions. Finally, the retention ratio
\begin{equation}
  \rho_l = \frac{R^2_l(U^A_l)}{R^2_l(I)}
  \label{eq:supp_rho}
\end{equation}
quantifies the fraction of total future information preserved within the action-sensitive subspace.

\paragraph{Quantitative findings.}
As summarized in Tab.~\ref{tab:supp_potent}, $\rho_l$ drops from an average of 0.77 in shallow layers (layers 0--3) to 0.62 in intermediate layers (layers 4--13) and 0.57 in deep layers (layers 14--29). Across 24 of the 30 layers, $R^2_l(U^A_l)$ falls below the 2.5th percentile of covariance-matched random subspaces, confirming that the action pathway systematically bypasses the feature directions carrying future dynamics.

\begin{table}[t]
  \centering
  \caption{Future decodability ($R^2_l(S)$, Eq.~\eqref{eq:supp_subspace_r2}) across 8-dimensional subspaces. \textit{Random} reports the median over 12 covariance-matched draws.}
  \label{tab:supp_potent}
  \small
  \setlength{\tabcolsep}{4pt}
  \begin{tabular}{lcccc}
    \toprule
    Layer & All features & Random & Action-sensitive & $\rho_l$ \\
    \midrule
    0  & 0.69 & 0.66 & 0.64 & 0.93 \\
    10 & 0.82 & 0.65 & 0.43 & 0.52 \\
    20 & 0.89 & 0.75 & 0.53 & 0.60 \\
    29 & 0.89 & 0.71 & 0.41 & 0.46 \\
    \bottomrule
  \end{tabular}
\end{table}

\section{Causal Erasure Protocol}
\label{sec:supp_erase}
\setcounter{equation}{0}
\renewcommand{\theequation}{D\arabic{equation}}

\paragraph{Residual future target and concept eraser.}
To evaluate causal dependence without directly ablating current proprioceptive or action cues, we extract the residual future displacement orthogonal to the ground-truth action chunk $a$ and robot state $s$:
\begin{equation}
  r = \Delta z - W^{\ast}[a; s],
  \label{eq:residual}
\end{equation}
where $W^{\ast}$ is fitted via ordinary least squares on the training episodes. For each layer $l$, we identify a subspace $U_l$ of rank 128--192 encoding $r$ via iterative nullspace projection~\cite{ravfogel2020null} on the analysis episodes (Appendix~\ref{sec:supp_setup}). During evaluation, visual tokens are projected onto the nullspace of $U_l$:
\begin{equation}
  \tilde{x} = x - U_l U_l^{\top}(x - \mu_l),
  \label{eq:erase}
\end{equation}
where $\mu_l$ is the layer feature mean. Retaining any overlap between $U_l$ and action-sensitive directions provides a conservative upper bound on the true action shift caused by future erasure.

\paragraph{Sliding-window intervention and metrics.}
We intervene across sliding windows of four consecutive layers $[l, l{+}3]$ ($l \in \{0,\dots,26\}$) by modifying only the cached visual keys and values passed to the Action DiT, leaving the Video DiT forward pass intact. As a control, we erase covariance-matched random subspaces of identical rank. Evaluation is performed on 100 disjoint episodes using 10 flow-matching denoising steps. For each episode $n$, we compare the erased action chunk $\tilde{a}_n$ against two intact chunks $\hat{a}_n$ and $\hat{a}'_n$ generated under independent noise seeds via the normalized action shift $\delta$ and ground-truth error ratio $\kappa$:
\begin{equation}
  \delta = \frac{\mathrm{median}_n\, \mathrm{RMS}(\tilde{a}_n - \hat{a}_n)}{\mathrm{median}_n\, \mathrm{RMS}(\hat{a}'_n - \hat{a}_n)}, \qquad
  \kappa = \frac{\sum_n \lVert \tilde{a}_n - a_n \rVert^2}{\sum_n \lVert \hat{a}_n - a_n \rVert^2},
  \label{eq:shift}
\end{equation}
with 95\% confidence intervals estimated via 1,000 paired bootstrap resamples.

\paragraph{Results.}
Tab.~\ref{tab:erasure} reports representative layer windows in Fast-WAM. Across all depths, erasing residual future cues yields action shifts ($\delta$) and error ratios ($\kappa$) nearly indistinguishable from random-subspace erasure, confirming that predictive representations in the shared backbone remain largely unutilized without an explicit relay interface.

\begin{table}[t]
  \centering
  \small
  \caption{Causal erasure across four-layer windows in Fast-WAM.}
  \label{tab:erasure}
  \begin{tabular}{lcccc}
    \toprule
    & \multicolumn{2}{c}{Action shift $\delta$} & \multicolumn{2}{c}{Error ratio $\kappa$} \\
    \cmidrule(lr){2-3}\cmidrule(lr){4-5}
    Layers & Future & Random & Future & Random \\
    \midrule
    8--11  & 0.23 & 0.17 & 1.17 & 1.01 \\
    14--17 & 0.36 & 0.27 & 1.35 & 1.34 \\
    18--21 & 0.44 & 0.25 & 1.11 & 1.09 \\
    22--25 & 0.41 & 0.39 & 1.20 & 1.11 \\
    26--29 & 1.06 & 0.90 & 1.28 & 1.21 \\
    \bottomrule
  \end{tabular}
\end{table}

\section{Implementation and Benchmark Details}
\label{sec:supp-exp-setup}

\paragraph{Architecture and training hyperparameters.}
RelayWAM builds upon the Wan2.2-5B Video DiT backbone~\cite{wan}, paired with its native text encoder and VAE. Guided by the layer-wise analysis in Sec.~\ref{sec:analysis}, recurrent transition layers are attached starting at layer $l_0 = 14$ through layer $29$, introducing 34.96M parameters ($<0.6\%$ of the 6.02B backbone). All models and controlled variants are trained for 20,000 steps on $8\times$ NVIDIA A100-80GB GPUs using AdamW with a constant learning rate of $1\times 10^{-4}$ and a global batch size of 128.

\paragraph{Evaluation protocols.}
For LIBERO~\cite{libero}, a single policy is jointly trained across the four standard suites (Spatial, Object, Goal, and Long; 10 tasks per suite) and evaluated over 50 rollouts per task (500 episodes per suite). Zero-shot out-of-distribution generalization is evaluated directly on LIBERO-Plus~\cite{liberoplus} (7 perturbation axes) and LIBERO-PRO~\cite{liberopro} (5 generalization axes) using official evaluation splits without additional fine-tuning.